\documentclass[runningheads]{llncs}
\usepackage{graphicx}
\usepackage{epstopdf}
\usepackage{amsmath,amssymb} 
\usepackage{color}

\usepackage{caption}
\usepackage{subcaption}
\usepackage{mathabx}
\usepackage{epsfig}
\usepackage{float}
\floatstyle{plaintop}
\restylefloat{table}

\newcommand{\ignore}[1]{}

\begin{document}

\newcommand{\point}{
    \raise0.7ex\hbox{.}
    }


\pagestyle{headings}

\mainmatter

\title{3D Shape Retrieval via Irrelevance Filtering and Similarity Ranking (IF/SR)} 

\titlerunning{3D Shape Retrieval via Irrelevance Filtering and Similarity Ranking (IF/SR)} 

\authorrunning{Xiaqing Pan, Yueru Chen, C.-C. Jay Kuo} 
\author{Xiaqing Pan, Yueru Chen, C.-C. Jay Kuo}
\institute{Ming-Hsieh Department of Electrical Engineering, University of Southern California, Los Angeles, CA 90089-2564, USA} 

\maketitle

\begin{abstract}
A novel solution for the content-based 3D shape retrieval problem using
an unsupervised clustering approach, which does not need any label
information of 3D shapes, is presented in this work.  The proposed shape
retrieval system consists of two modules in cascade: the irrelevance
filtering (IF) module and the similarity ranking (SR) module.  The IF
module attempts to cluster gallery shapes that are similar to each other
by examining global and local features simultaneously.  However, shapes
that are close in the local feature space can be distant in the global
feature space, and vice versa. To resolve this issue, we propose a joint
cost function that strikes a balance between two distances.  Irrelevant
samples that are close in the local feature space but distant in the
global feature space can be removed in this stage.  The remaining
gallery samples are ranked in the SR module using the local feature.
The superior performance of the proposed IF/SR method is demonstrated by
extensive experiments conducted on the popular SHREC12 dataset. 
\end{abstract}

\section{Introduction} \label{sec.introduction}

Content-based 3D shape retrieval \cite{tangelder2008survey} has received
a lot of attention in recent years due to an rapidly increasing number
of 3D models over the Internet (e.g., Google sketchup and Yobi3D).
Applications of 3D shape retrieval technologies include: 3D model
repository management, mechanical components retrieval, medical organ
model analysis, etc.  Given a 3D shape model as the query, a
content-based 3D shape retrieval system analyzes the query shape and
retrieves ranked 3D shapes from the gallery set according to a
similarity measure. Its performance is evaluated by consistency between
ranked shapes and human interpretation.  A robust and efficient 3D shape
retrieval system is needed for users to access and exploit large 3D
datasets effectively. 

Recently, convolutional neural-network (CNN) based solutions achieved
impressive performance by training a network using either multiple views
of 3D shapes \cite{su2015multi,xie2015deepshape,shi2015deeppano,savvashrec,Bai_2016_CVPR} or the 3D volumetric data \cite{wu20153d,maturana2015voxnet,qi2016volumetric}. However, their
training procedure demands a large amount of labeled data, which is
labor-intensive. In this work, we address the 3D shape retrieval problem
using an unsupervised learning approach. It has broader applications
since no labeled data are needed. 

The main challenge in 3D shape retrieval lies in a wide range of shape
variations. A generic 3D shape dataset such as SHREC12 \cite{li2012shrec} includes both
rigid and non-rigid shapes. Shape variations can be categorized into
inter-class similarities and intra-class variations. For the latter, we
have articulation, surface deformation, noise, etc.  

Global and/or local features can be used to measure the similarity
between two 3D shapes. The rotation invariant spherical harmonics (RISH)
\cite{kazhdan2003rotation} and the D2 shape distribution
\cite{osada2002shape} are two representative global features. They
capture object surface properties using the frequency decomposition and
the vertex distance histogram, respectively.  The retrieval performance
using global features only may degrade due to the loss of fine shape
details. To overcome this limitation, research efforts in recent years
have focused on developing more discriminative local features. They can
be categorized into surface-based and view-based features. Surface-based
local features \cite{smeets2013meshsift,bronstein2010scale,bronstein2011shape,gal2007pose,reuter2006laplace} describe a local surface region to achieve pose oblivion, scale and orientation invariance.  Although surface-based
retrieval methods are effective in handling non-rigid shape retrieval \cite{lian2011shape,lian2015shrec},
they are not robust against shape artifacts that do occur in generic
shape datasets. Retrieval methods using view-based local features are
favored for this reason. 

\begin{figure}[!t]
\centering
\begin{subfigure}[b] {0.9\textwidth}
\centering
\includegraphics[width=\textwidth]{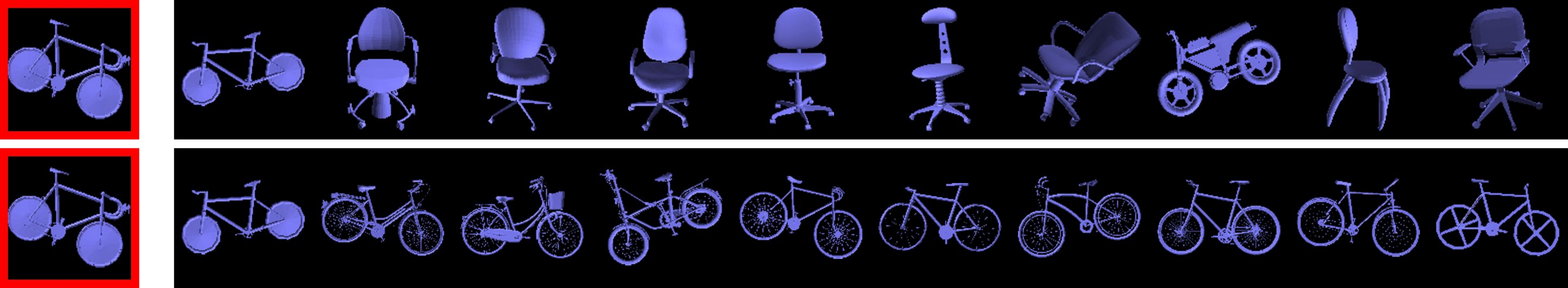}
\caption{Bicycle}
\end{subfigure}
\begin{subfigure}[b]{0.9\textwidth}
\centering
\includegraphics[width=\textwidth]{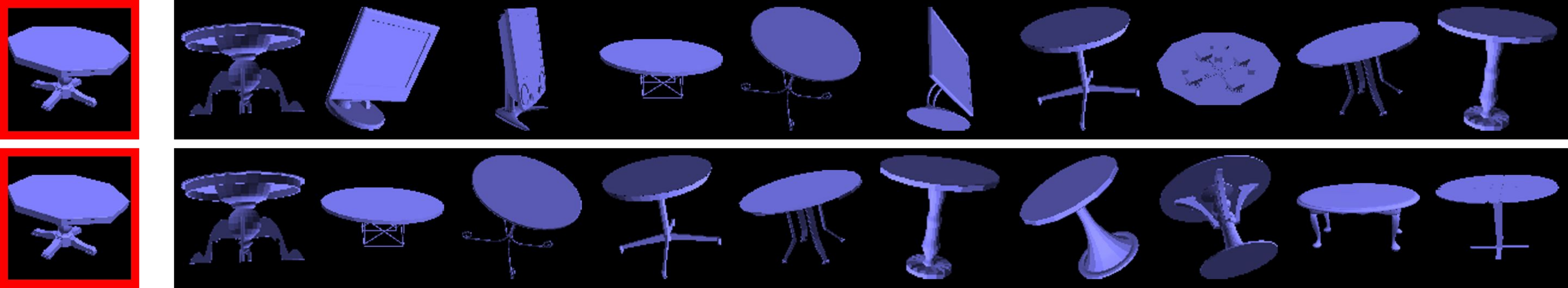}
\caption{Round table}
\end{subfigure}
\begin{subfigure}[b]{0.9\textwidth}
\centering
\includegraphics[width=\textwidth]{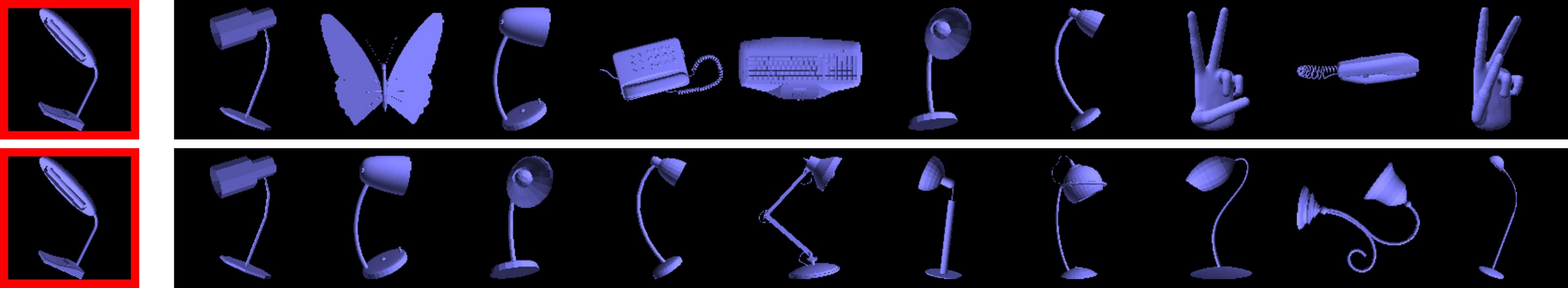}
\caption{Desk lamp}
\end{subfigure}
\begin{subfigure}[b] {0.9\textwidth}
\centering
\includegraphics[width=\textwidth]{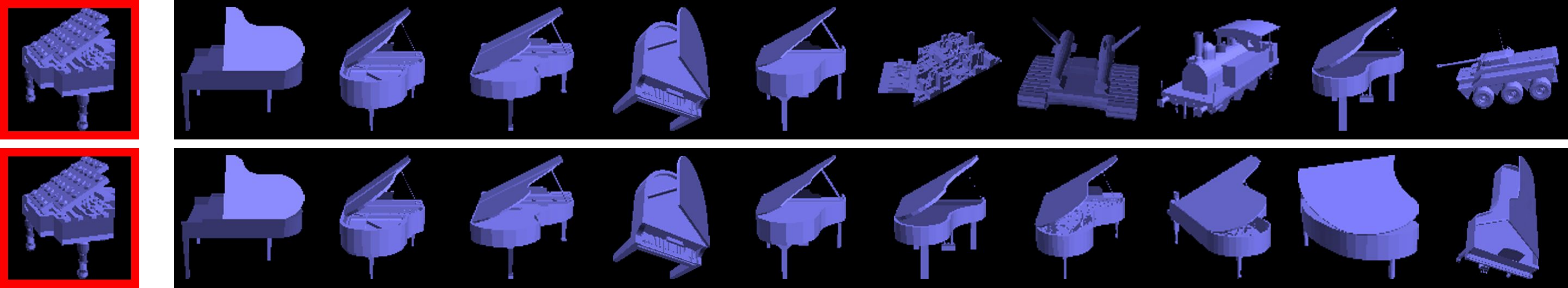}
\caption{Piano}
\end{subfigure}
\begin{subfigure}[b]{0.9\textwidth}
\centering
\includegraphics[width=\textwidth]{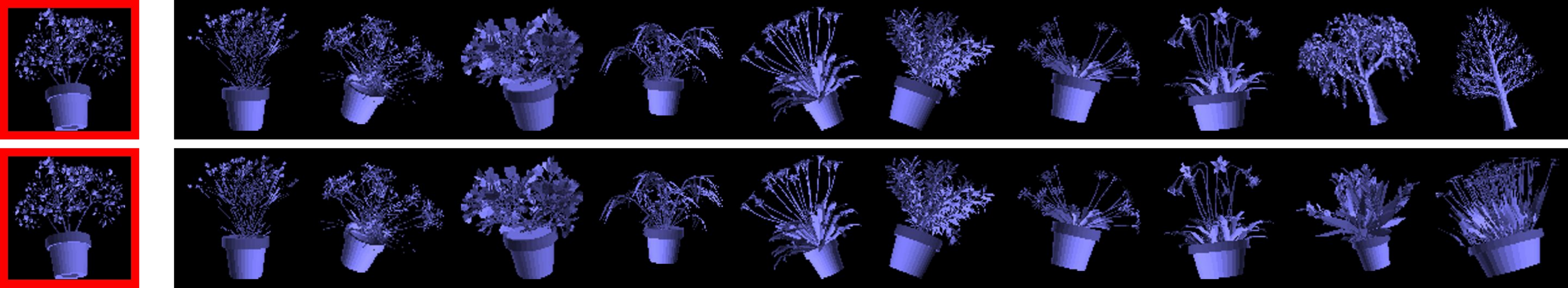}
\caption{Home plant}
\end{subfigure}
\caption{Comparison of retrieved shapes using the DG1SIFT method (the
first row) and the proposed IF/SR method (the second row) against five
query shapes (from top to bottom): (a) bicycle, (b) round table, (c)
desk lamp, (d) piano, and (e) home plant. }\label{fig.slvf_tsr_compare}
\end{figure}

View-based methods project a 3D shape into multiple views. Generally
speaking, an adequate number of view samples can represent a 3D shape
well.  The light field descriptor (LFD) method \cite{chen2003visual} and
the multi-view depth line approach (MDLA) \cite{chaouch2007new}
represent each view by Zernike moments plus polar Fourier descriptors
and depth lines, respectively. The similarity between two shapes is
measured by enumerating multiple rotation setups. The salient local
visual feature (SLVF) method \cite{ohbuchi2008salient} extracts the SIFT
points \cite{lowe1999object} from each range view. After constructing a
codebook from the training pool, one feature of a 3D shape can be
represented by the histogram of SIFT points from all views using the Bag
of Words (BoW) approach. The DG1SIFT method \cite{ohbuchi2010distance}
extends the SLVF method by extracting three types of SIFT points from
each view.  They are dense, global and one SIFTs. The depth buffered
super vector coding (DBSVC) method \cite{li2014shrec} uses a dense power
SURF feature and the super vector coding algorithm to improve the
feature discriminability.  

Although local features achieve a better performance than global
features when being tested against several generic 3D shape datasets
\cite{li2012shrec,bronstein2010shrec,dutagaci2011shrec,li2015comparison},
their discriminative power is restricted in the global scale. In
particular, they may retrieve globally irrelevant 3D shapes in high
ranks.  To illustrate this point, we show five retrieval results by
applying the DG1SIFT method to the SHREC12 3D shape dataset in Fig.
\ref{fig.slvf_tsr_compare} (a)-(e).  With the five query shapes in the
leftmost column, the top 10 retrieved results are presented in the first
row of each subfigure Obviously, errors in these retrieval results are
counter to human intuition.  Being motivated by the observation, we propose
a more robust 3D shape retrieval system which is called the irrelevance
filtering and similarity ranking (IF/SR) method. Its retrieved results
are shown in the second row of each subfigure.  All mistakes generated
by DG1SIFT are corrected by our method. Clearly, the proposed IF/SR
system has a more robust performance as shown in these examples. 

There are two main contributions of this work.  First, we develop more
powerful and robust global features to compensate for the weaknesses of
local features. Feature concatenation are often adopted by traditional
methods to combine local and global features.  However, proper feature
weighting and dimension reduction remain to be a problem.  For the
second contribution, we propose a robust shape retrieval system that
consists of two modules in cascade: the irrelevance filtering (IF)
module and the similarity ranking (SR) module.  The IF module attempts
to cluster gallery shapes that are similar to each other by examining
global and local features simultaneously.  However, shapes that are
close in the local feature space can be distant in the global feature
space, and vice versa. To resolve this issue, we propose a joint cost
function that strikes a balance between two distances. In particular,
irrelevant samples that are close in the local feature space but distant
in the global feature space can be removed in this stage.  The remaining
gallery samples are ranked in the SR module using the local feature. 

The rest of this paper is organized as follows. The proposed IF/SR
method is explained in Section \ref{sec.3dretrieval_TSR}. Experimental
results are shown in Section \ref{sec.experiment}. Finally, concluding
remarks are given in Section \ref{sec.conclusion}. 

\section{Proposed IF/SR Method} \label{sec.3dretrieval_TSR}

\subsection{System Overview} \label{subsec.3Dtsroverview}

\begin{figure*}[!t]
\begin{center}
\includegraphics[width=.9\textwidth]{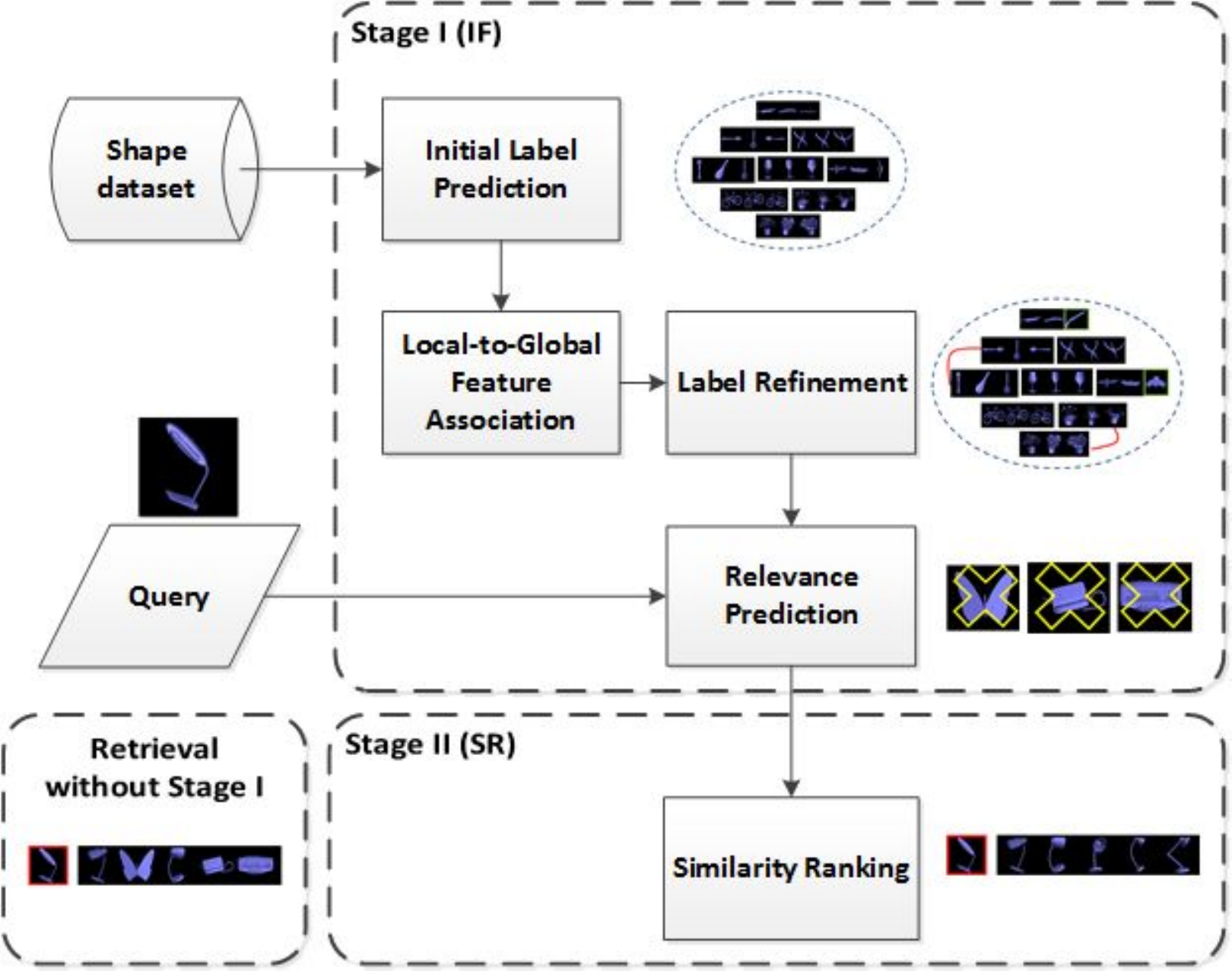}
\caption{The flow chart of the proposed IF/SR system.}\label{fig.flowchart3Dshape}
\end{center}
\end{figure*}

The flow chart of the proposed IF/SR method is shown in Fig.
\ref{fig.flowchart3Dshape}. The IF module is trained in an off-line
process with the following three steps. 
\begin{enumerate}
\itemsep -1ex
\item {\bf Initial label prediction}. All gallery samples are assigned
an initial label in their local feature space using an unsupervised
clustering method. \\
\item {\bf Local-to-global feature association}. Samples close to each
cluster centroid are selected as the training data. A random forest
classifier is trained based on their global features. All gallery
samples are re-predicted by the random forest classifier to build an
association from the local feature space to the global feature space. \\
\item {\bf Label refinement.} We assign every gallery sample a set of
relevant cluster indices based on a joint cost function. The joint cost
function consists of two assignment scores. One score reflects the
relevant cluster distribution of the query sample itself while the other
is the mean of the relevant cluster distributions of its local
neighbors. The ultimate relevant cluster indices are obtained by
thresholding the cost function. 
\end{enumerate}

In the on-line query process, we extract both global and local features
from a query shape and proceed with the following two steps. 
\begin{enumerate}
\itemsep -1ex
\item {\bf Relevance prediction.} we adopt the same scheme in the label
refinement step to assign relevant cluster indices to a given query. \\
\item {\bf Similarity ranking.} The similarity between the query and all
relevant gallery samples is measured in the local feature space. In this
step, a post-processing technique can also be adopted to enhance retrieval
accuracy. 
\end{enumerate}

An exemplary query, desk lamp, is given in Fig.
\ref{fig.flowchart3Dshape} to illustrate the on-line retrieval process.
In the dashed box ``retrieval without Stage I", the traditional local
feature (DG1SIFT) retrieves erroneous shapes such as butterfly, desk
phone and keyboard in the top five ranks. They are apparently irrelevant
to the query shape and successfully removed in the relevance prediction
step in the IF stage (Stage I). The retrieved top five samples in the SR
stage (Stage II) are all desk lamp shapes. We will explain the
processing of Stages I and II in detail below. 

\subsection{Stage I: Irrelevance Filtering} \label{subsec.3Dpreprocessing}

{\bf 3D Shape Preprocessing}. We have two preprocessing steps: 1) model
representation conversion and 2) 3D shape normalization. For model
representation conversion, since we extract features from both mesh
models and volumetric models, we adopt the parity count method
\cite{nooruddin2003simplification} to convert a mesh model into a
volumetric model. Each volumetric model has resolution $256\times 256
\times 256$. 3D shape normalization aims to align shapes of the same
class consistently to achieve translational, scaling and rotational
invariance. 

\begin{figure}[!t]
\centering
\begin{subfigure}[b]{.45\textwidth}
\centering
\includegraphics[width=\textwidth]{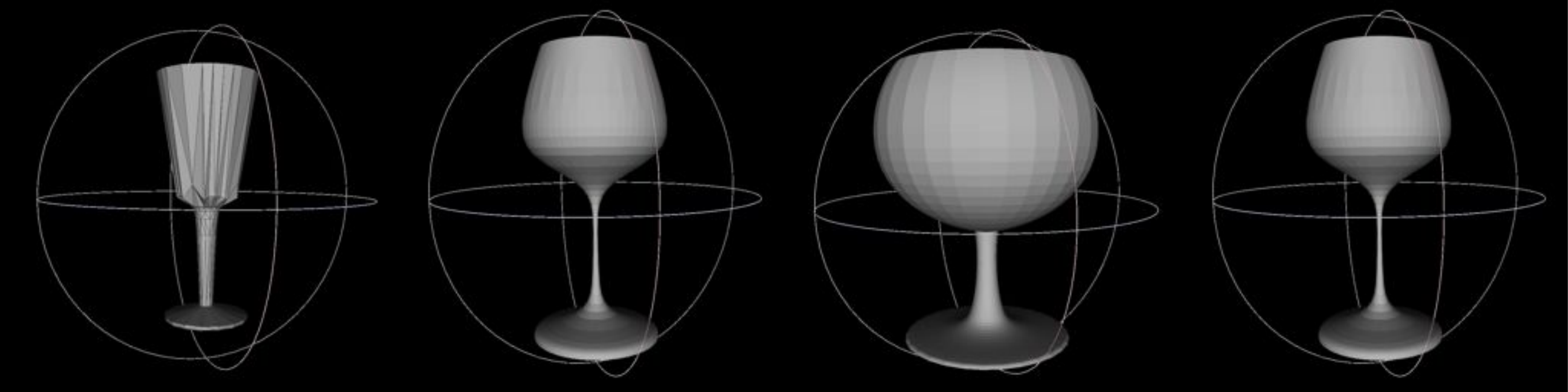}
\caption{Cup}
\end{subfigure}
\begin{subfigure}[b]{.45\textwidth}
\centering
\includegraphics[width=\textwidth]{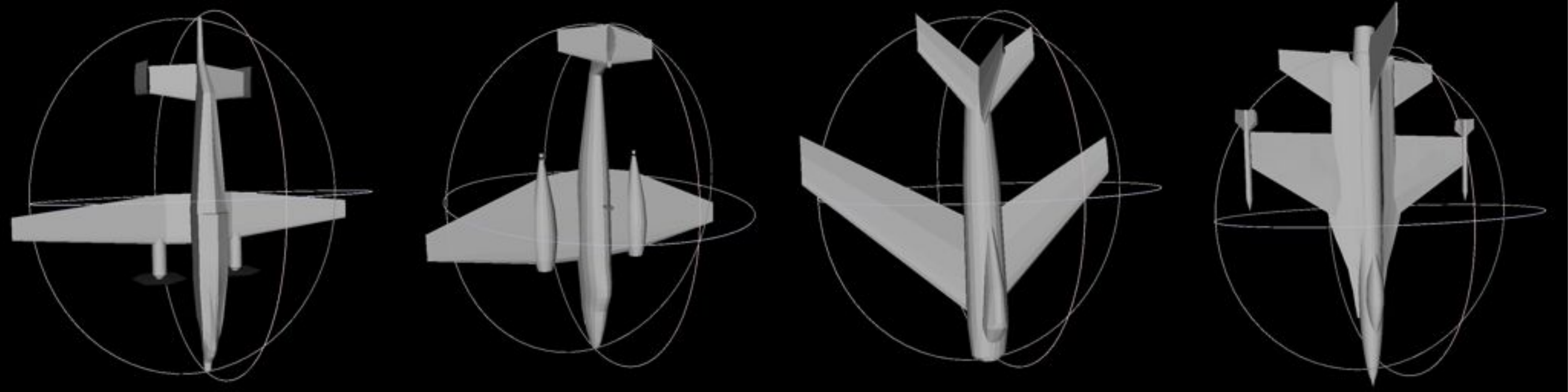}
\caption{Monoplane}
\end{subfigure}
\begin{subfigure}[b]{.45\textwidth}
\centering
\includegraphics[width=\textwidth]{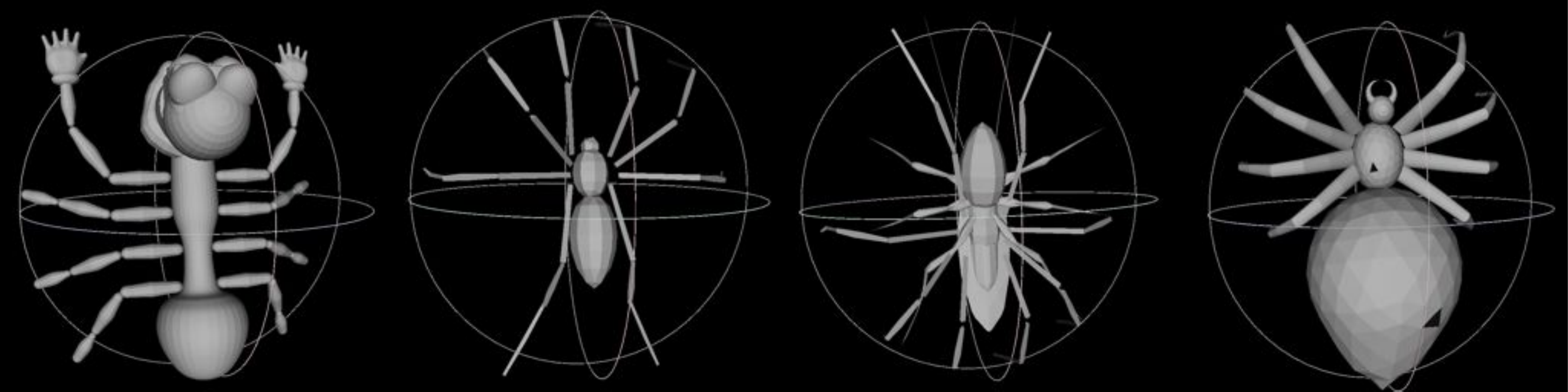}
\caption{Non-flying insect}
\end{subfigure}
\begin{subfigure}[b]{.45\textwidth}
\centering
\includegraphics[width=\textwidth]{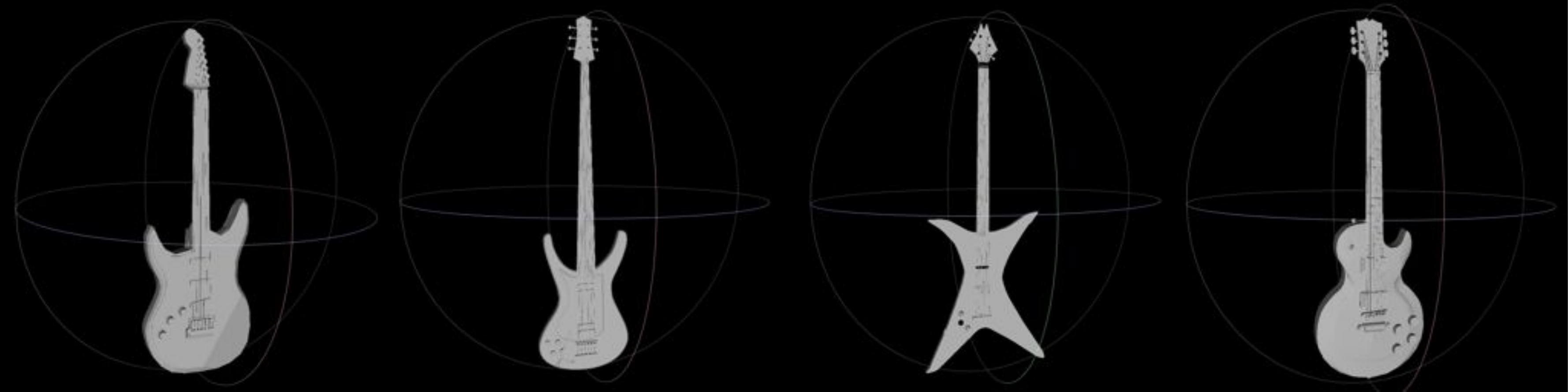}
\caption{Guitar}
\end{subfigure}
\caption{Shape normalization results of four 3D shape classes.}\label{fig.sym_nol_res}
\end{figure}

Translational invariance is achieved by aligning the center of mass with
the origin.  For scale invariance, we re-scale a shape to fit a unit
sphere. For rotational invariance, we adopt the reflective symmetry
axial descriptor \cite{kazhdan2004reflective} to calculate the nearly
complete symmetry function for each 3D shape. The PCA on the symmetry
function extracts three principal axes to form three principal planes.
To determine the order of three principal planes, we project the shape
into each plane and the projection views with the first and second
largest areas are aligned with the XOY plane and the ZOX plane,
respectively.  Finally, the YOZ plane is determined automatically.  Fig.
\ref{fig.sym_nol_res} shows some normalization results using the
above-mentioned method. 

{\bf Global Features}. To capture the global properties of a 3D shape,
we describe it using three feature types: 1) surface features ($f_s$),
2) wavelet features ($f_w$) and 3) geometrical features ($f_g$). 

\begin{figure}[!t]
\centering
\begin{subfigure}[b] {0.45\textwidth}
\centering
\includegraphics[width=\textwidth]{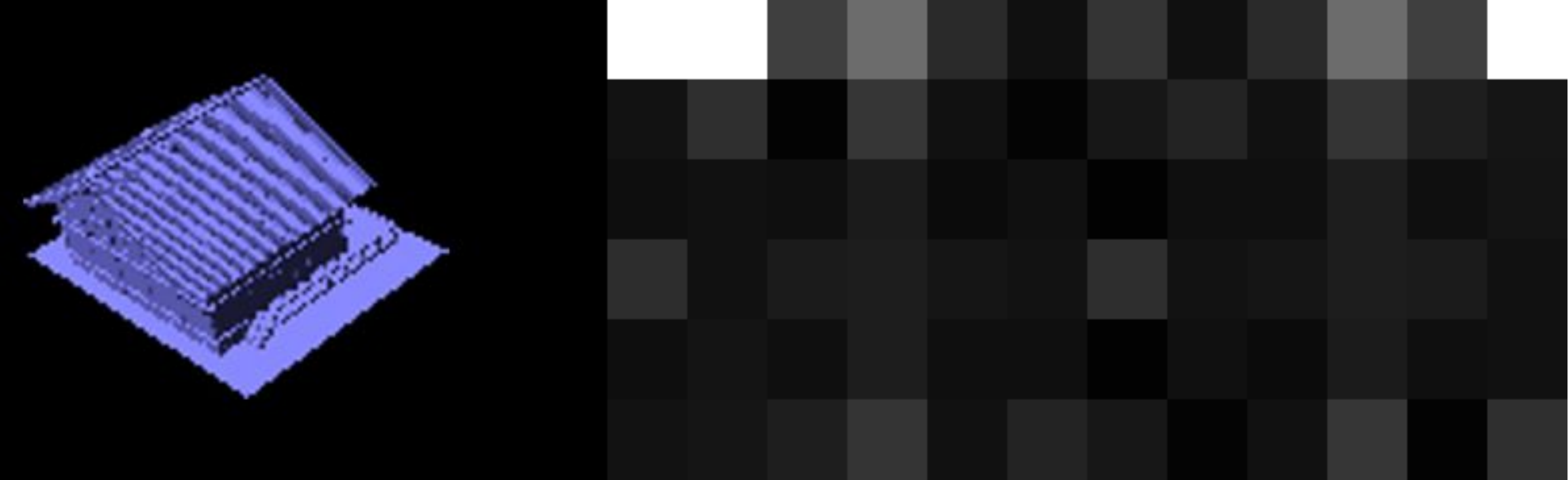}
\caption{}
\end{subfigure}
\begin{subfigure}[b]{0.45\textwidth}
\centering
\includegraphics[width=\textwidth]{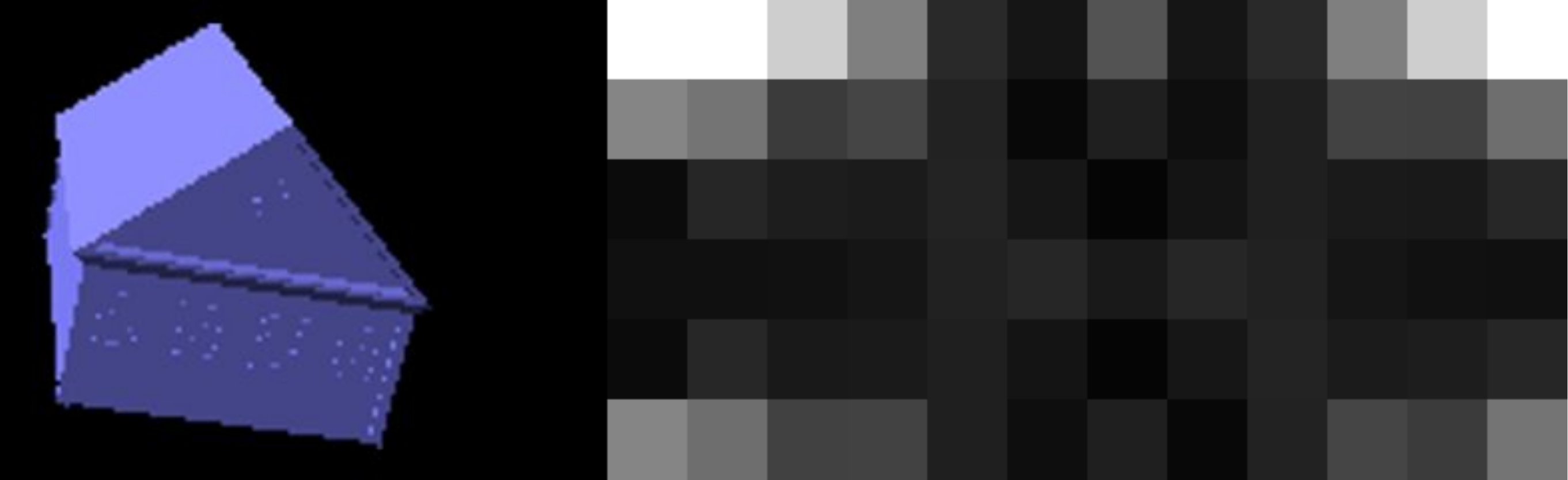}
\caption{}
\end{subfigure}
\begin{subfigure}[b]{0.45\textwidth}
\centering
\includegraphics[width=\textwidth]{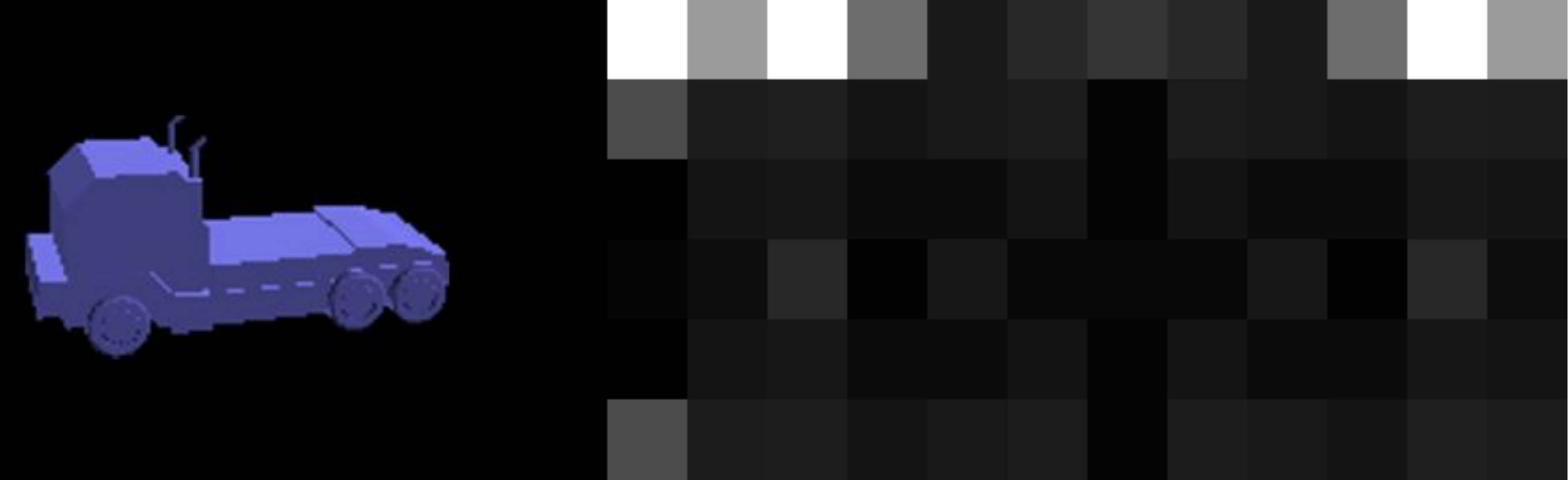}
\caption{}
\end{subfigure}
\begin{subfigure}[b]{0.45\textwidth}
\centering
\includegraphics[width=\textwidth]{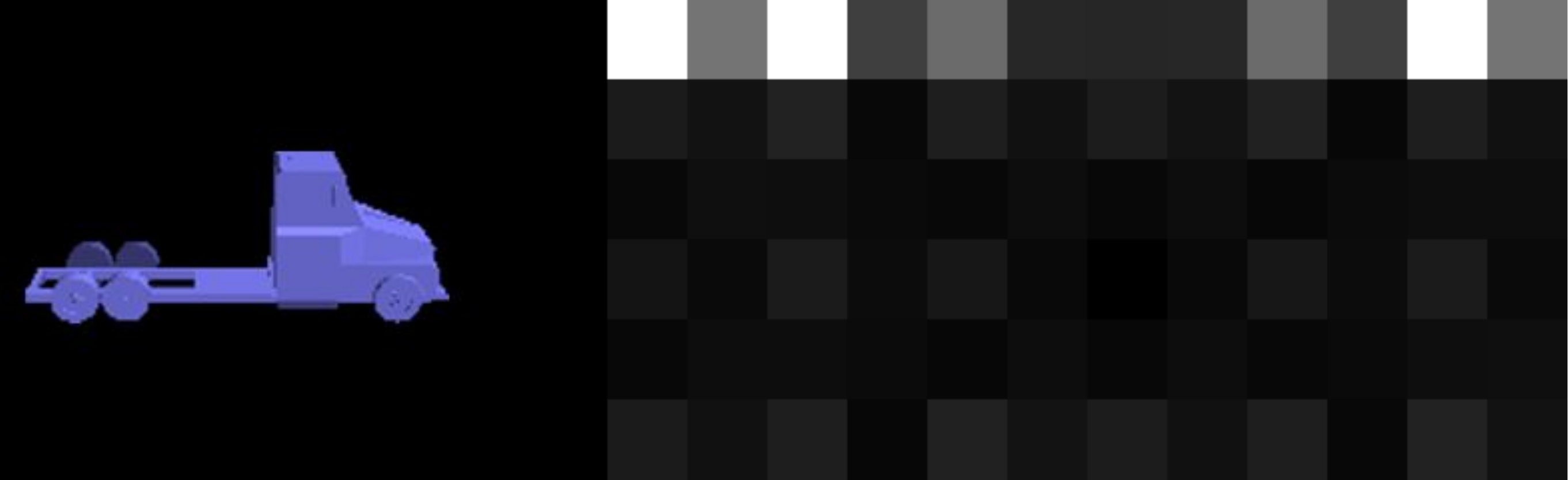}
\caption{}
\end{subfigure}
\caption{Visualization of surface features of four shapes, where (a) 
and (b) provide two house shapes while (c) and (d) provide two truck
shapes.}\label{fig.dh_map}
\end{figure}

The 3D surface features, denoted by $f_s$, are a generalization of the
2D polar Fourier descriptor \cite{zhang2002integrated}. $N$ rays are
emitted from the origin of a normalized 3D shape. Each ray has the
orientation $\mathbf{r} = (cos\phi cos\theta, cos\phi sin \theta, sin
\phi)$ with two directional parameters $(\theta, \phi)$, where $\theta$
and $\phi$ are uniformly sampled from intervals $[0, \pi)$ and $[0,
2\pi)$, respectively, with step size $\frac{\pi}{6}$.  For each ray, the
Euclidean distance from the origin to its intersected point on a face
forms a function $g(\theta, \phi)$. If a ray intersects with multiple
faces, we consider the farthest one only. In this way, we convert the
original surface function $f(x, y, z)$ into a 2D distance function
parameterized by $g(\theta, \phi)$. Then, we calculate the Fourier
coefficients of the 2D distance function.  The magnitude information
forms a 72-D feature vector denoted by $f_s$. The Fourier descriptors of
four shapes belonging to two classes are visualized in Fig.
\ref{fig.dh_map}, where each subfigure contains an original shape in the
left and its surface feature in the right. We see intra-class
consistency and inter-class discrimination from this figure. 

\begin{figure}[!th]
\centering
\includegraphics[width=\textwidth]{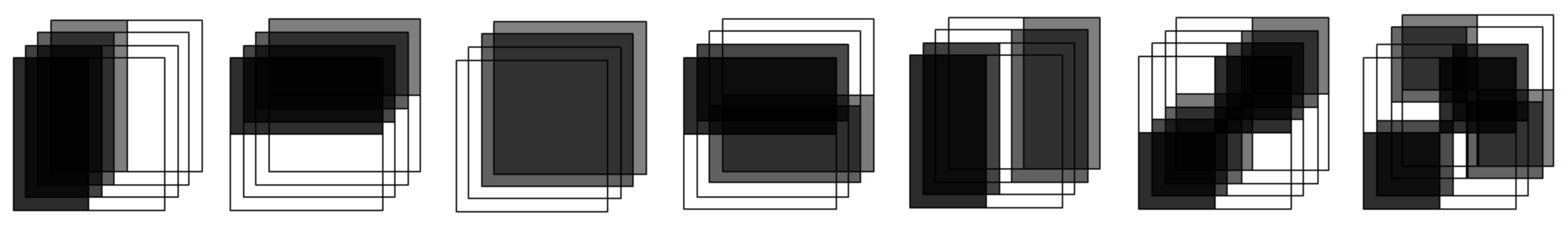}
\caption{Illustration of the seven-band Haar filters.}\label{fig.haar_ex}
\end{figure}

For the wavelet features denoted by $f_w$, we adopt the generalized 3D
Haar-like filters \cite{cui20073d}. Seven bands of 3D Haar-like filters
as shown in Fig.  \ref{fig.haar_ex} are applied to a normalized and
voxelized model. The first three filters capture the left-right,
top-bottom, front-back symmetry properties. The last four filters
analyze diagonal sub-regions.  The responses from these seven filters
form a 7D wavelet feature vector. 

\begin{figure}[!th]
\centering
\includegraphics[width=\textwidth]{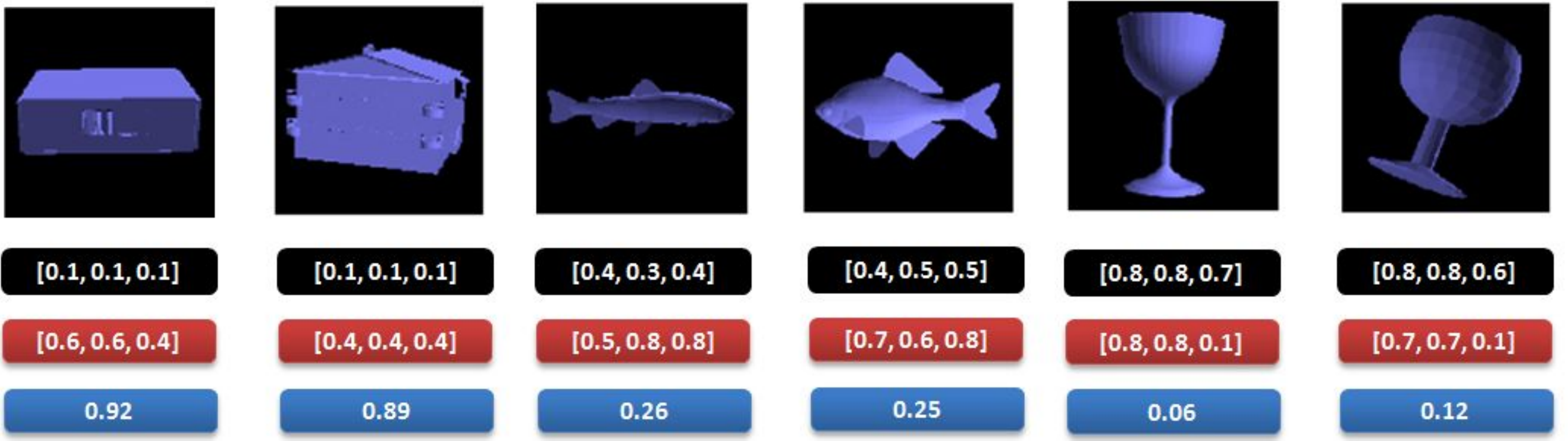}
\caption{The xyz-invariance (black box), $\alpha\beta\gamma$-variance
(red box) and rectilinearity (blue box) values of six examples from
three classes: apartment house, fish and cup.}\label{fig.rect_ex}
\end{figure} 

Furthermore, we incorporate four geometrical features: 1) the aspect
ratio, 2) xyz-invariance, 3) $\alpha\beta\gamma$-invariance and 4)
rectilinearity\cite{lian2010rectilinearity}. The aspect ratio is a 3D
feature based on three side lengths - $l_x$, $l_y$, $l_z$ of the
bounding box of a normalized shape. It is expressed as
\begin{equation}
AR = [\frac{l_x}{l_x+l_y+l_z}, \frac{l_y}{l_x+l_y+l_z}, \frac{l_z}{l_x+l_y+l_z}]. 
\end{equation}
The xyz-variance and $\alpha\beta\gamma$-variance are adopted to examine
the variance of cut-planes of a normalized volumetric model. To measure
the xyz-variance, we extract all cut-planes orthogonal to the X-axis,
the Y-axis and the Z-axis, respectively. The variances of three groups
of cut-planes form a 3D feature. Similarly, the
$\alpha\beta\gamma$-variance measures the variance of groups of rotated
cut-planes centered at the X-axis, the Y-axis and the Z-axis,
repectively.  The robust rectilinearity measure from
\cite{lian2010rectilinearity} is used to obtain the rectilinearity
feature. It calculates the ratio between the total surface area and the
sum of projected triangle areas on the XOY, the ZOX and the YOZ planes.
Finally, the geometrical feature, denoted by $f_g$, is a 10-D feature
vector. The geometric features of six examples are shown in Fig.
\ref{fig.rect_ex} in boxes of black (xyz-invariance), red
($\alpha\beta\gamma$-variance) and blue (rectilinearity), respectively. 

{\bf Initial Label Prediction.} In traditional 3D shape retrieval
formulation, all shapes in the dataset are not labeled. Under this
extreme case, we select the spectral clustering algorithm
\cite{ng2002spectral} to reveal the underlying relationship between
gallery samples. The local feature is strong at grouping locally similar
shapes but it is sensitive to local variances as discussed in Section
\ref{sec.introduction}. In contrast, the global feature is powerful at
differentiating global dissimilar shapes but weak at finding locally
similar shapes. Thus, the combination of the two in this early stage
tends to cause confusion and lower the performance. For this reason, we
use the local feature only to perform clustering. 

\begin{figure}[!t]
\centering
\includegraphics[width=.95\textwidth]{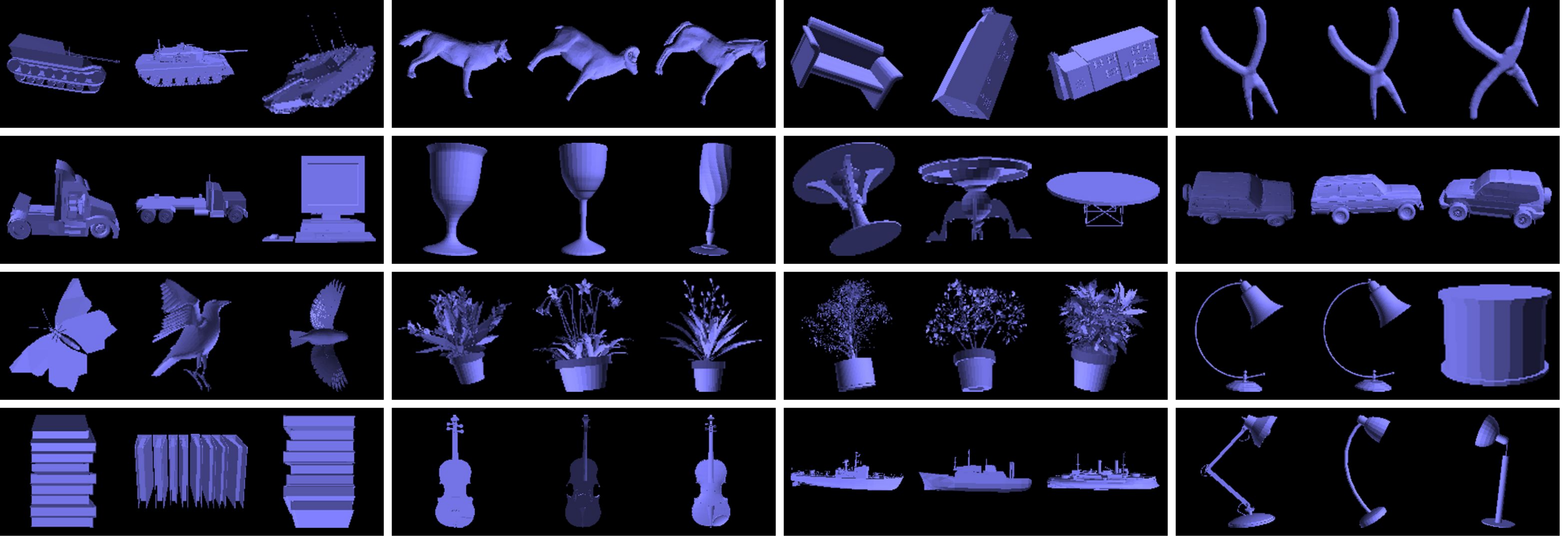}
\caption{Several clusterd SHREC12 shapes using the spectral clustering
method using the DG1SIFT feature.}\label{fig.cluster_ex}
\end{figure}

For the SHREC12 dataset, shapes in several clusters using the DG1SIFT
feature are shown in Fig. \ref{fig.cluster_ex}. Some clusters look
reasonable while others do not. Actually, any unsupervised clustering
method will encounter two challenges. First, uncertainty occurs near
cluster boundaries so that samples near boundaries have a higher
probability of being wrongly clustered. Second, the total number of
shape classes is unknown. When the cluster number is larger than the
class number in the database, the clustering algorithm creates
sub-classes or even mixed classes. We address the first challenge in the
local-to-global feature association step and the second challenge in the
label refinement step. 

{\bf Local-to-Global Feature Association}. We extract $N_k$ samples
closest to the centroid of the $k^{th}$ cluster and assign them a
cluster label. Clearly, samples sharing the same cluster label are close
to each other in the feature space. There is a trade-off in choosing a
proper value of $N_k$. A smaller $N_k$ guarantees higher clustering
accuracy but fewer gallery samples will be assigned cluster labels.
Empirically, we set the value of $N_k$ to one half of the size of the
$k^{th}$ cluster. Then, we convert the gallery samples from the local
feature space to a global feature space. We will correct clustering
errors in the global feature space at a later stage. Furthermore,
samples that come from the same class but are separated in the local
feature space can be merged by their global features. To build the
association, labeled samples are used to train a random forest
classifier \cite{breiman2001random} with global features. Finally, all
gallery shapes are treated as test samples. The random forest classifier
is used to predict the probability of each cluster type by voting. In
this way, samples clustered in the local feature space can be linked to
multiple clusters probabilistically due to the similarity in the global
feature space. 

\begin{figure}[!th]
\centering
\begin{subfigure}[b] {0.7\textwidth}
\centering
\includegraphics[width=\textwidth]{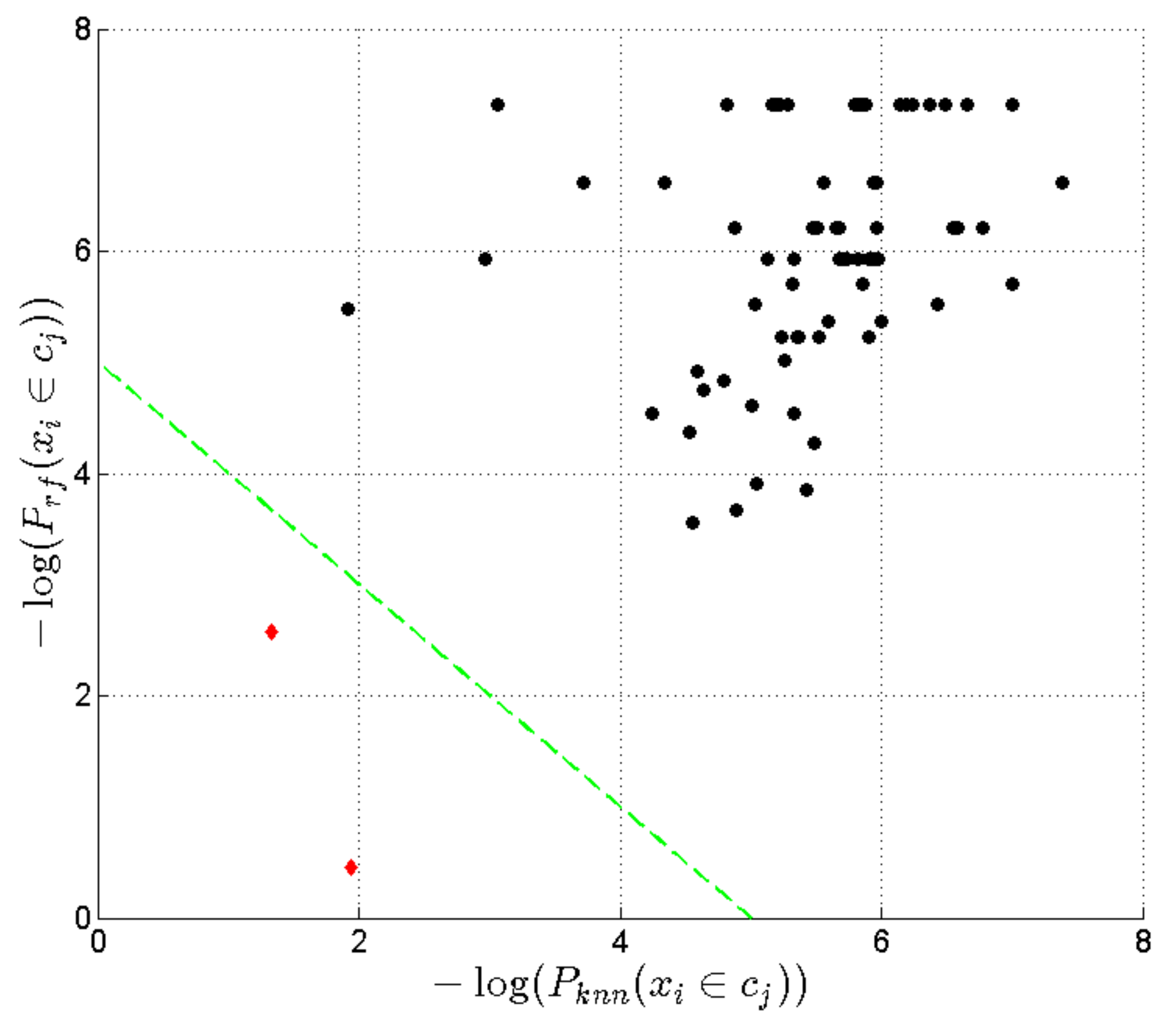}
\caption{}
\end{subfigure}
\begin{subfigure}[b]{0.7\textwidth}
\centering
\includegraphics[width=\textwidth]{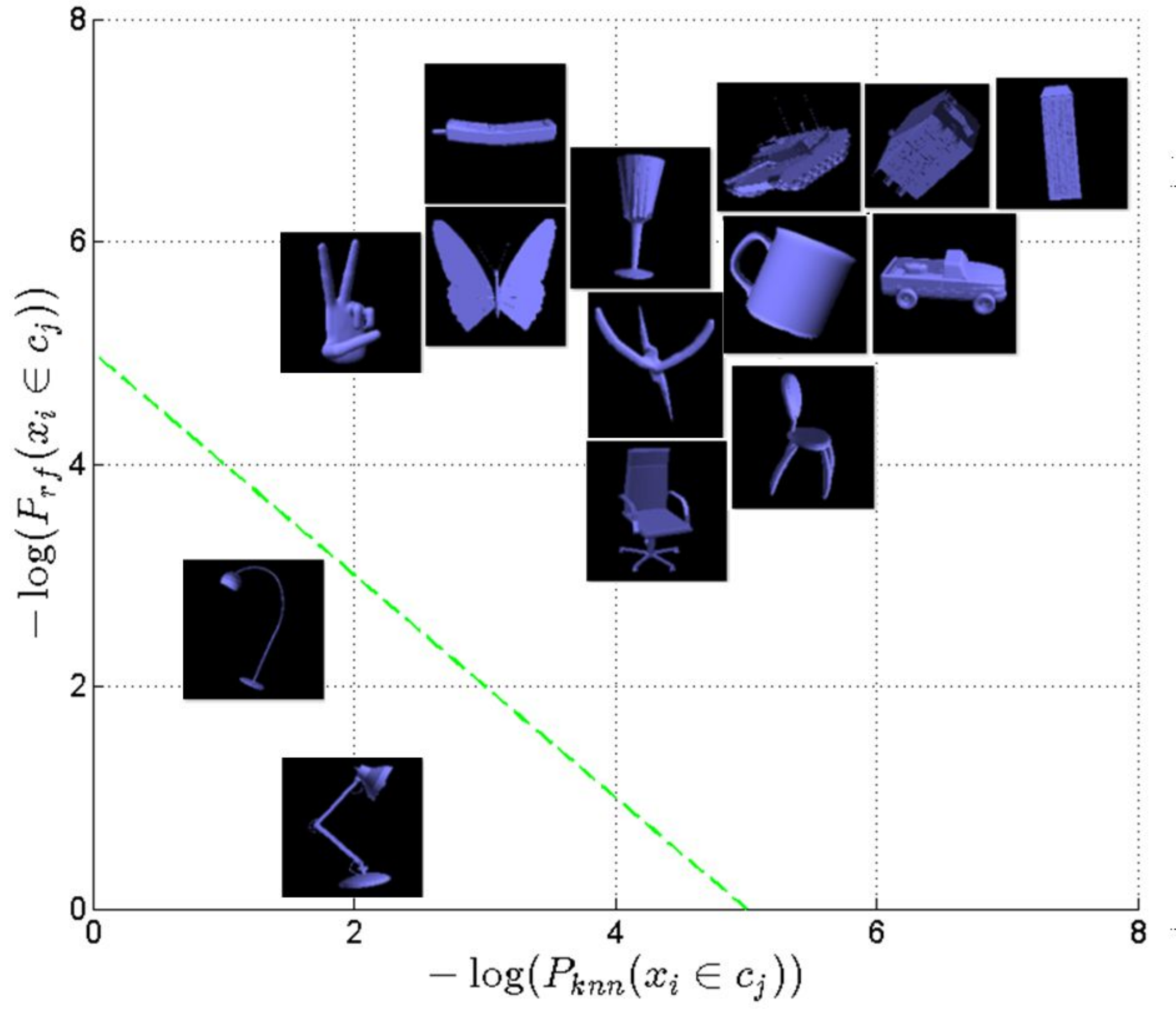}
\caption{}
\end{subfigure}
\caption{Selecting relevant clusters for the query desk lamp in Fig.
\ref{fig.slvf_tsr_compare}(c) by thresholding a cost function shown
in Eq. \ref{eq.cost}. }\label{fig.cost}
\end{figure}

{\bf Label Refinement}. The output of the IF module includes: 1) a set of
indexed clusters, and 2) soft classification (or multi-labeling) of all
gallery samples. For item \#1, we use unsupervised spectral clustering
to generate clusters as described above.  If the class number is known
(or can be estimated), it is desired that the cluster number is larger
than the class number. Each of these clusters is indexed by a cluster
ID.  For item \#2, we adopt soft classification so that each sample can
be associated with multiple clusters. This is done for two reasons. If
two sub-classes belong to the same ground truth class, we need a
mechanism to re-group them together.  Clearly, a hard classification
process does not allow this to happen.  Second, a hard classification
error cannot be easily compensated while a soft classification error is
not as fatal and it is likely to be fixed in the SR module (stage II). 

We consider two relevant cluster assignment schemes below.

1) \textit{Direct Assignment} 

We apply the random forest classifier to both training and testing
samples based on their global features. Then, the probability for the
$i^{th}$ shape sample (denoted by $y_i$) belonging to the $k^{th}$ cluster
(denoted by $c_k$) can be estimated by the following normalized voting result:
\begin{equation} \label{eq.rf_prob}
P_{rf}(y_i \in c_k) = \frac{v_k}{\sum_j{v_j}},
\end{equation}
where $v_k$ is the number of votes claiming that $y_i$ belongs to $c_k$.
Eq. (\ref{eq.rf_prob}) associates $y_i$ to its relevant clusters directly. 

2) \textit{Indirect Assignment} \\

Intuitively, a good cluster relevance assignment scheme should take both
global and local features into account.  For query sample, $y_i$, we
find its $K$ nearest neighbors (denoted by $x_j$) using a certain
distance function in a local feature space (e.g.  the same feature space
used in DG1SIFT).  Then, the probability of $y_i$ belonging to $c_k$ can
be estimated by the weighted sum of the probability in Eq.
(\ref{eq.rf_prob}) in form of
\begin{equation} \label{eq.knn_prob}
P_{knn}(y_i \in c_k) =\! \frac{\sum_{x_j \in knn(y_i)}{P_{rf}(x_j
\in {c_k})}}{\sum_{c_m}{\sum_{x_j \in knn(y_i)}{P_{rf}(x_j \in {c_m})}}}.
\end{equation}
Eq.  (\ref{eq.knn_prob}) associates $y_i$ to its relevant clusters
indirectly. That is, the assignment is obtained by averaging the
relevant clusters assignment of its $K$ nearest neighbors. Empirically,
we choose K to be 1.5 times the average cluster size \ignore{$K=20$} in
the experiments. 

We show an example that assigns a query desk lamp shape to its relevant
clusters in Fig. \ref{fig.cost}(a), whose x-axis and y-axis are the
negative log functions of Eqs. (\ref{eq.rf_prob}) and
(\ref{eq.knn_prob}), respectively.  Every dot in Fig. \ref{fig.cost}(a)
represents a cluster after shape clustering. To visualize shapes
represented by a dot, we plot a representative sample of each cluster in
Fig. \ref{fig.cost}(b). 

We see that the distance between the hand cluster and the desk lamp
cluster is small in the x-axis but large in the y-axis. This is because
that samples of the desk lamp and hand clusters are interleaved in the
local feature space as shown in the retrieval results of DG1SIFT in Fig.
\ref{fig.slvf_tsr_compare}(c).  However, the desk lamp and the hand
clusters have little intersection in the global feature space.  In
contrast, the wheel chair and desk lamp clusters have large intersection
in the global feature space. Yet, their distance is far in the local
feature space. It is apparent that Eqs. (\ref{eq.rf_prob}) and
(\ref{eq.knn_prob}) provide complementary relevance assignment
strategies for query sample $y_i$. It is best to integrate the two into
one assignment scheme. For example, we can draw a line to separate
relevant and irrelevant clusters with respect to the query apple shape
in this plot. 

Mathematically, we can define the following cost function 
\begin{align} \label{eq.cost}
\mathbb{J}(y_i,c_k) &= - \log(P_{knn}(y_i \in c_k) P_{rf}(y_i \in c_k)) \nonumber \\
    &= - [\log(P_{knn}(y_i \in c_k)) + \log(P_{rf}(y_i \in c_k))].
\end{align}
We compute $\mathbb{J}(y_i,c_k)$ for all clusters $c_k$. If
\begin{equation}\label{eq.thresholding}
\mathbb{J}(y_i,c_k)<\epsilon,
\end{equation}
where $\epsilon$ is a pre-selected threshold. We say that cluster $c_k$
is a relevant cluster for query $y_i$. Otherwise, it is irrelevant. 
 
\subsection{Stage II: Similarity Ranking}

In the SR module, we rank the similarity between a given query and
gallery samples in the retrieved relevant clusters using a
local-features-based matching scheme (e.g., DG1SIFT). Additionally, we
adopt the Local Constrained Diffusion Process (LCDP)
\cite{yang2009locally} in the post-processing step. The diffusion
process is slightly modified with the availability of relevant clusters
in the IF/SR system since the diffusion process can be conducted on a
more reasonable manifold due to the processing in Stage I. 

\section{Experimental Results} \label{sec.experiment}

We demonstrate the retrieval performance of the proposed IF/SR method by
conducting experiments on the generic 3D shape dataset of
SHREC12 \cite{li2012shrec}. It contains 1200 3D shapes in 60 independent
classes. Samples are uniformly distributed so that each class has 20
shape samples. The retrieval performance is measured by five standard
metrics. They are: Nearest-Neighbor (NN), First-Tier score (FT),
Second-Tier score (ST), E-measurement (E), Discounted Cumulative Gain
(DCG). 

We compare the proposed IF/SR method with five state-of-the-art methods:
\begin{itemize}
\itemsep -1ex
\item LSD-sum \cite{bai2012software}. It uses a local surface-based feature
that considers local geodesic distance distribution and Bag-of-Words.
\item ZFDR \cite{li20133d}. It adopts a hybrid feature that integrates 
the Zernike moment, the Fourier descriptor, the ray-based features. 
\item 3DSP\_L2\_1000\_chi2 \cite{li2012shrec}. It employs a local
surface-based feature that computes the 3D SURF descriptor under the
spatial pyramid matching scheme. 
\item DVD+DB+GMR \cite{li2012shrec}. It adopts a hybrid feature that
contains a dense voxel spectrum descriptor and a depth-buffer shape
descriptor. 
\item DG1SIFT \cite{ohbuchi2010distance}. It uses a view-based feature
that extracts three types of SIFT features (Dense SIFT, Grid SIFT and
One SIFT) per view. 
\end{itemize}

The IF/SR method adopts DG1SIFT as the local feature for shape
clustering.  We show the first-tier (FT) scores of the IF/SR method using a different
cluster number $M$ for shape clustering in Table
\ref{table.score_on_M_3D}.  Generally speaking, the performance degrades
when $M$ is small due to the loss of discriminability in larger cluster
sizes. The retrieval performance improves as the cluster number
increases up to 64.  After that, the performance saturates and could
even drop slightly. That means that we lose the advantage of clustering
when the cluster size is too small. For the remaining experimental
results, we choose $M=64$. 

\begin{table*}[!t]
\begin{center}
\resizebox{0.6\textwidth}{!}{
\begin{tabular}{|c|c|c|c|c|c|c|c|}
\hline
M          & 16      & 32      & 48      & 64               & 80      & 96     & 112      \\ \hline
FT         & 0.666   & 0.672   & 0.709   & \textbf{0.720}   & 0.717   & 0.717  & 0.715  \\ \hline
\end{tabular}}
\end{center}
\caption{Comparison of the First-Tier (FT) scores with different cluster numbers
for the IF/SR method in the SHREC12 dataset. The best score is shown in bold.} 
\label{table.score_on_M_3D}
\end{table*}

\begin{table}[!t]
\begin{center}
\resizebox{0.7\textwidth}{!}{
\begin{tabular}{|c|c|c|c|c|c|}
\hline
Method               & NN             & FT             & ST             & E              & DCG            \\ \hline
LSD-sum              & 0.517          & 0.232          & 0.327          & 0.224          & 0.565          \\ \hline
ZFDR                 & 0.818          & 0.491          & 0.621          & 0.442          & 0.776          \\ \hline
3DSP\_L2\_1000\_chi2 & 0.662          & 0.367          & 0.496          & 0.346          & 0.678          \\ \hline
DVD+DB+GMR           & 0.828          & 0.613          & 0.739          & 0.527          & 0.833          \\ \hline
DG1SIFT              & 0.879          & 0.661          & 0.799          & 0.576          & 0.871          \\ \hline
IF/SR                & \textbf{0.896}          & 0.720          & 0.837          & 0.608          & 0.891          \\ \hline
IF/SR+LCDP              & 0.893 & \textbf{0.734} & \textbf{0.858} & \textbf{0.620} & \textbf{0.899} \\ \hline
\end{tabular}}
\end{center}
\caption{Comparison of the NN, FT, ST, E and DCG scores of five
state-of-the-art methods, the proposed IF/SR method, and the IF/SR
method with LCDP postprocessing for the SHREC12 dataset. The best score
for each measurement is shown in bold.}
\label{table.SHREC12sixmeasurements}

\end{table}
 
\begin{table}[!th]
\begin{center}
\resizebox{.7\textwidth}{!}{
\begin{tabular}{|c|c|c|c|c|c|}
\hline
N                    & 20             & 25             & 30             & 35              & 40            \\ \hline
LSD-sum              & 0.232          & 0.260          & 0.286          & 0.310          & 0.327          \\ \hline
ZFDR                 & 0.491          & 0.539          & 0.575          & 0.603          & 0.621          \\ \hline
3DSP\_L2\_1000\_chi2 & 0.367          & 0.411          & 0.446          & 0.476          & 0.496          \\ \hline
DVD+DB+GMR           & 0.613          & 0.656          & 0.691          & 0.719          & 0.739          \\ \hline
DG1SIFT              & 0.661          & 0.718          & 0.756          & 0.783          & 0.799          \\ \hline
IF/SR                  & 0.720          & 0.775          & 0.802          & 0.824          & 0.837          \\ \hline
IF/SR+LCDP             & \textbf{0.734}           & \textbf{0.786}          & \textbf{0.817}          & \textbf{0.841}          & \textbf{0.858}          \\ \hline
\end{tabular}}
\end{center}
\caption{Comparison of top 20, 25, 30, 35, 40 retrieval accuracy for the
SHREC12 dataset, where the best results are shown in bold.}
\label{table.SHREC12topk}
\end{table}

\begin{figure}[!h]
\centering
\begin{subfigure}[b] {0.95\textwidth}
\centering
\includegraphics[width=\textwidth]{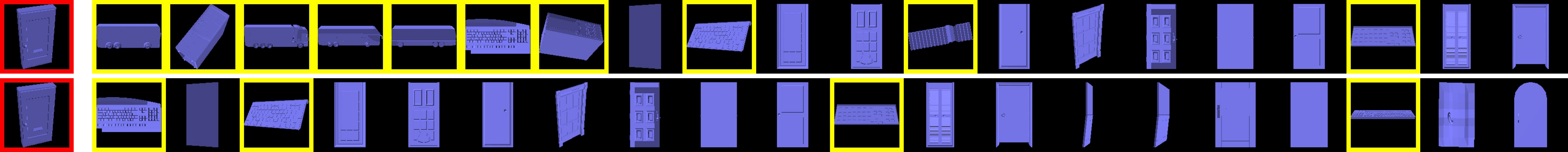}
\caption{Door}
\end{subfigure}
\begin{subfigure}[b]{0.95\textwidth}
\centering
\includegraphics[width=\textwidth]{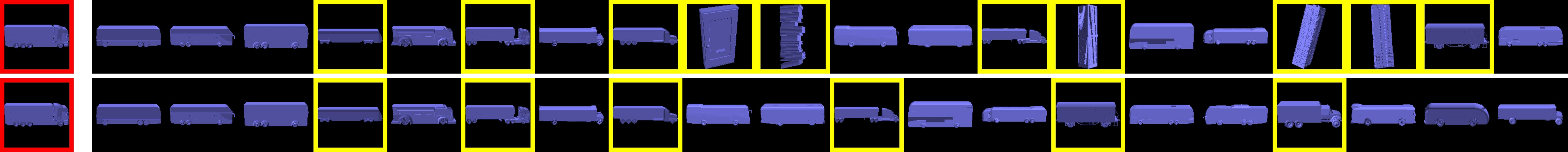}
\caption{Bus}
\end{subfigure}
\begin{subfigure}[b]{0.95\textwidth}
\centering
\includegraphics[width=\textwidth]{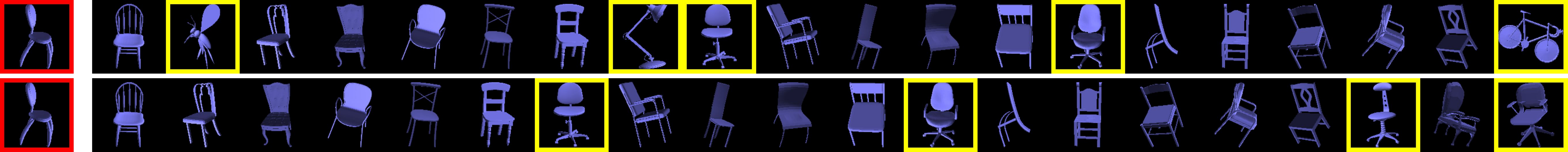}
\caption{Non-wheel chair}
\end{subfigure}
\begin{subfigure}[b]{0.95\textwidth}
\centering
\includegraphics[width=\textwidth]{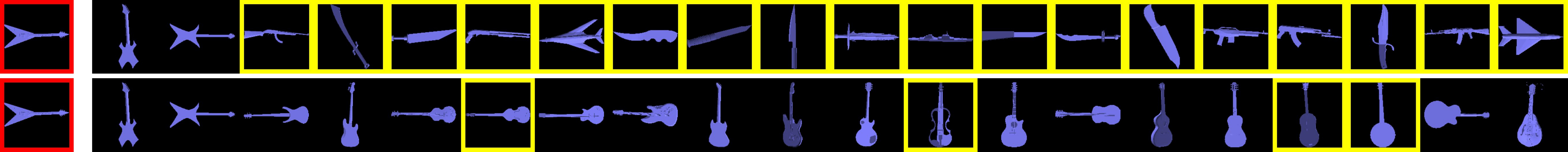}
\caption{Guitar}
\end{subfigure}
\begin{subfigure}[b]{0.95\textwidth}
\centering
\includegraphics[width=\textwidth]{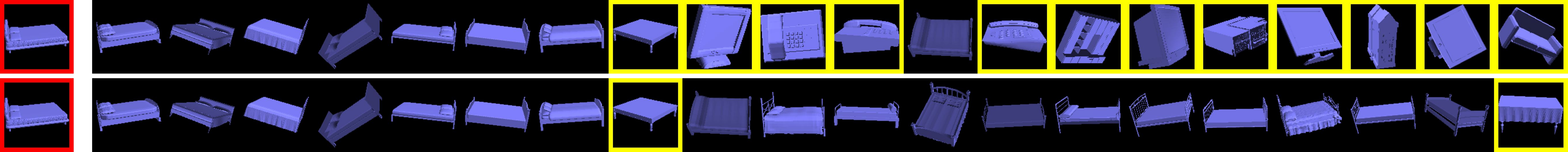}
\caption{Bed}
\end{subfigure}
\caption{Comparison of retrieved top 20 rank-ordered shapes. For each
query case given in the leftmost column, retrieved results of DG1SIFT
and the proposed IF/SR method are shown in the first and second rows of
all subfigures, repectively.}\label{fig.slvf_tsr_compare_err}
\end{figure}
 
We compare the performance of seven 3D shape retrieval methods with five
measures in Table \ref{table.SHREC12sixmeasurements}.  Clearly, the
proposed IF/SR method (with or without LCDP postprocessing) outperforms
the other five benchmarking methods. The IF/SR method with
postprocessing improves the result of DG1SIFT by around $7\%$ in the
First-Tier score. Since DG1SIFT adopts the manifold ranking process in
its similarity measurement, the gap between the IF/SR method before and
after LCDP is relatively small. 

Since each SHREC12 shape class contains 20 shape samples, the measure of
correctly retrieved samples from the top 20 (FT) and 40 (ST) ranks
cannot reflect the true power of the proposed IF/SR method. To push the
retrieval performance further, we compare the accuracy of retrieved
results from the top 20, 25, 30, 35 and 40 ranks of the IF/SR method and
five benchmarking methods in Table \ref{table.SHREC12topk}, whose
first and last columns correspond the FT and ST scores reported in Table
\ref{table.SHREC12sixmeasurements}.  The superiority of the IF/SR method
stands out clearly in this table. 
 
According to the top 20 retrieval performance, the IF/SR method still
makes mistakes for some queries. We conduct error analysis and show the
results of DG1SIFT and the IF/SR method in Figs.
\ref{fig.slvf_tsr_compare_err}(a)-(e).  For each query case given in the
leftmost column, retrieved results of DG1SIFT and the IF/SR method are
shown in the first and second rows of all subfigures, respectively. Each
erroneous result is enclosed by a thick frame. The errors of DG1SIFT are
obvious. They are far away from human experience.  The IF/SR method
makes mistakes between door/keyboard, bus/truck, non-wheel chair/wheel
chair, guitar/violin and bed/rectangle table (see the second row of all
subfigures).  These mistakes are more excusable since they are closer
to each other based on human judgment. 

\begin{figure}[!t]
\centering
\includegraphics[width = 0.85\textwidth]{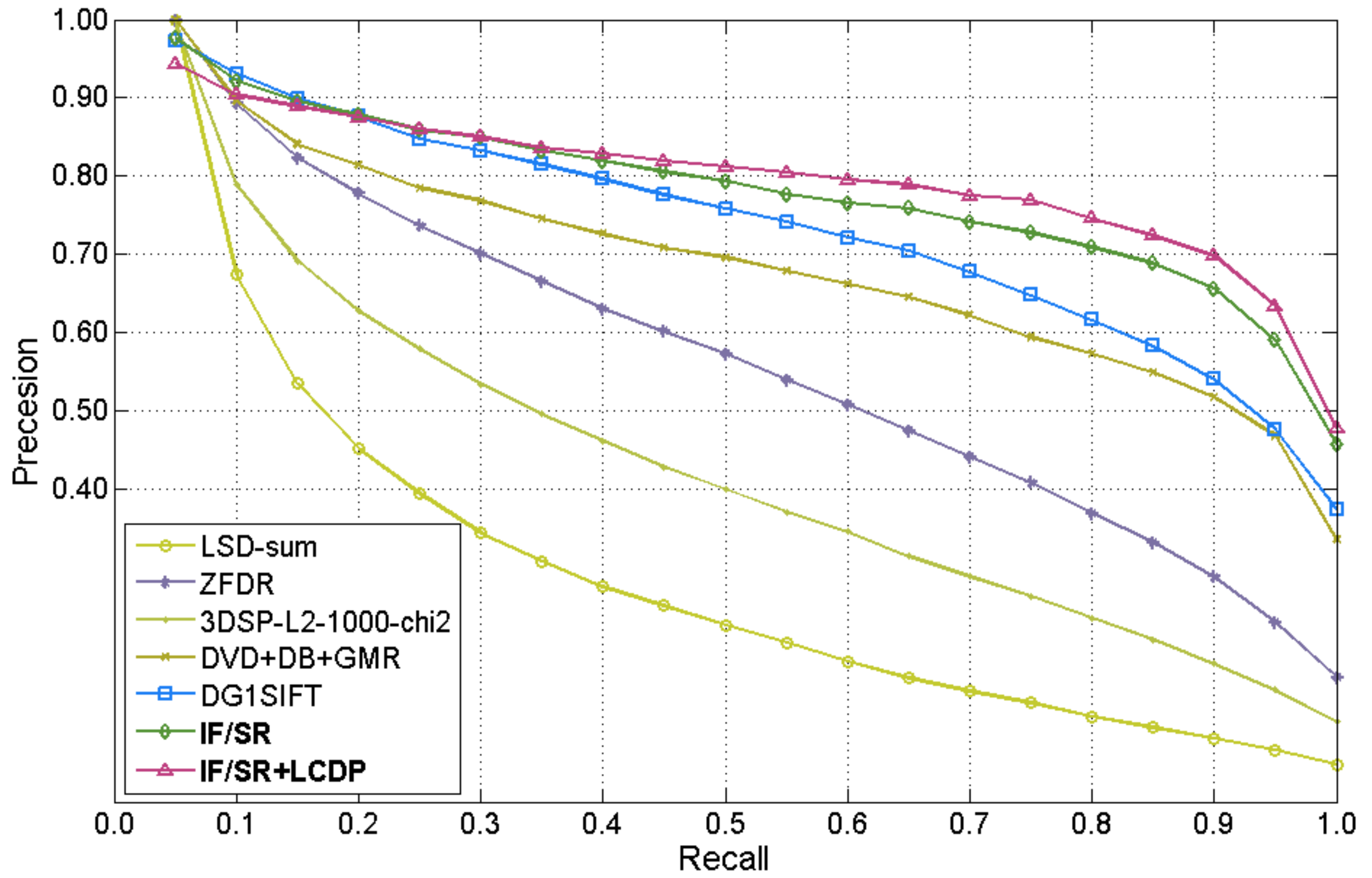}
\caption{Comparison of precision and recall curves of the proposed IF/SR method 
and several benchmarking methods for the SHREC12 dataset.}\label{fig.pr_shrec12}
\end{figure}

Finally, we show the precision-and-recall curves of the IF/SR method and
several methods in Fig.  \ref{fig.pr_shrec12}.  We see from the figure
that the IF/SR method outperforms all other methods by a significant
margin. 

\section{Conclusion} \label{sec.conclusion}

The IR/SF method was proposed to solve the unsupervised 3D shape
retrieval problem. In the IF stage, irrelevant shape clusters
are removed for each query shape. In the SR stage, the system
can focus on the matching and ranking in a much smaller subset of
shapes. It superior retrieval performance was evaluated on the
popular SHREC12 dataset.

\bibliographystyle{splncs}
\bibliography{ACCV_xiaqing}



\end{document}